\documentclass[11pt]{article}   	
\usepackage[margin=0.8in]{geometry}                		
\usepackage{tipa} 
\usepackage{indentfirst}
\usepackage{graphicx}				
								
\usepackage{amssymb}
\usepackage{color}
\usepackage{subfigure}
\usepackage{float}
\usepackage{amsmath}
\usepackage{booktabs}
\usepackage[flushleft]{threeparttable}
\usepackage{color}
\usepackage{sidecap}
\usepackage{xpatch}
\usepackage[font=small,labelfont=bf]{caption}
\usepackage{mathpazo} 
\usepackage[scaled]{helvet} 
\usepackage{courier} 
\normalfont
\usepackage[T1]{fontenc}
\usepackage{siunitx} 

\long\def\symbolfootnote[#1]#2{\begingroup%
\def\thefootnote{\fnsymbol{footnote}}\footnote[#1]{#2}\endgroup}
\def\blfootnote{\xdef\@thefnmark{}\@footnotetext}

\usepackage{graphicx}				
\usepackage{amssymb}
\usepackage{color}
\usepackage{subfigure}
\usepackage{amsmath}
\usepackage{booktabs}
\usepackage{mathrsfs}
\usepackage[flushleft]{threeparttable}
\usepackage{sidecap}
\usepackage{amssymb,bm,mathrsfs}
\usepackage{amsfonts, amsmath}
\usepackage{mathpazo} 
\usepackage[scaled]{helvet} 
\usepackage{courier} 

\normalfont
\usepackage[T1]{fontenc}
\usepackage{paralist}
\usepackage{multirow}

\usepackage{url}
\usepackage[hyperfootnotes=false]{hyperref}
\usepackage[semicolon,sort&compress]{natbib}

\usepackage{algorithm}
\usepackage{eqnarray}
\usepackage[noend]{algpseudocode}

\renewcommand{\cite}[1]{\citep{#1}}

\newcommand{\red}[1]{{\color{black} #1}}

\title{Simulated Annealing for Optimization of Graphs and Sequences}

\author{
\footnotesize{Xianggen Liu$^{1,3,4,\dag,\ddag}$,
	Pengyong Li$^{2,3,4, \dag}$,
	Fandong Meng$^5$, 
	Hao Zhou$^6$,   
	Huasong Zhong$^7$,
    Jie Zhou$^5$, } \\
    \footnotesize{Lili Mou$^8$,
	Sen Song$^{3,4,*}$}
	\\ 
\footnotesize{	$^1$College of Computer Science, Sichuan University, Chengdu 610065, China} \\
	\footnotesize{$^2$School of Computer Science and Technology, Xidian University, Xi’an 710071, China} \\
	\footnotesize{$^3$Laboratory for Brain and Intelligence and Department of Biomedical Engineering, Tsinghua University, Beijing 100084, China} \\
	\footnotesize{$^4$Beijing Innovation Center for Future Chip, Tsinghua University, Beijing 100084, China} \\
	\footnotesize{$^5$Pattern Recognition Center, WeChat AI, Tencent Inc, Beijing 100084, China}\\
	\footnotesize{$^6$ByteDance AI Lab, Beijing 100098, China}\\
	\footnotesize{$^7$MMU KuaiShou Inc, Beijing 100085, China}\\
	\footnotesize{$^8$Department of Computing Science, University of Alberta;}\\
	\footnotesize{Alberta Machine Intelligent Institute (Amii), Edmonton T6G 2R3, Canada}\\
	\footnotesize{\{liuxg16, lpy15\}@mails.tsinghua.edu.cn, \{fandongmeng, withtomzhou\}@tencent.com, zhouhao.nlp@bytedance.com,}\\ \footnotesize{\{zhonghsuestc,doublepower.mou\}@gmail.com,  songsen@mail.tsinghua.edu.cn}
}

\date{}	

\begin{document}

\maketitle
\symbolfootnote[0]{\footnotesize{$^\dag$} These authors contributed equally to this work.}
\symbolfootnote[0]{\footnotesize{$^\ddag$} This paper is an extension to \citet{upsa}, published at ACL 2020. There is more than 40\% new material, including graph optimization algorithms and several experiments on molecule generation. The code of our work is available at: https://github.com/liuxg16/UPSA}
\symbolfootnote[0]{\footnotesize{$^*$} To whom correspondence should be addressed. Email:  \protect\url{songsen@mail.tsinghua.edu.cn}}
\symbolfootnote[0]{\footnotesize{$^\star$} This article is an accepted manuscript of \textit{Neurocomputing} and under the CC-BY-NC-ND license. The formal publication of this manuscript is: Xianggen Liu, Pengyong Li, Fandong Meng, Hao Zhou, Huasong Zhong, Jie Zhou, Lili Mou, Sen Song. (2021). Simulated annealing for optimization of graphs and sequences. \textit{Neurocomputing}, 465:310-324. https://doi.org/10.1016/j.neucom.2021.09.003}
\begin{abstract}
Optimization of discrete structures aims at generating a new structure with the better property given an existing one, which is a fundamental problem in machine learning. Different from the continuous optimization, the realistic applications of discrete optimization (e.g., text generation) are very challenging due to the complex and long-range constraints, including both syntax and semantics, in discrete structures. In this work, we present SAGS, a novel Simulated Annealing framework for Graph and Sequence optimization. The key idea is to integrate powerful neural networks into metaheuristics (e.g., simulated annealing, SA) to restrict the search space in discrete optimization. We start by defining a sophisticated objective function, involving the property of interest and pre-defined constraints (e.g., grammar validity). SAGS searches from the discrete space towards this objective by performing a sequence of local edits, where deep generative neural networks propose the editing content and thus can control the quality of editing. We evaluate SAGS on paraphrase generation and molecule generation for sequence optimization and graph optimization, respectively. Extensive results show that our approach achieves state-of-the-art performance compared with existing paraphrase generation methods in terms of both automatic and human evaluations. Further, SAGS also significantly outperforms all the previous methods in molecule generation.$^{\star}$
\end{abstract}


\newpage
\section{Introduction}
Materials that the world consists of are mainly discrete and structured, from the quanta to stars and from neural spikes to human languages. Therefore, learning to manipulate discrete structured data is a fundamental problem in machine learning. As an example, the optimization of discrete structured data (e.g., graphs and sequences) involves generating a new sample with desirable properties, which relates to a variety of applications such as drug discovery~\citep{silva2019novo,xiao2013optimization} and text generation~\citep{kulkarni2013babytalk,wiseman-rush-2016-sequence}.

However, optimization on discrete structures is challenging. On the one hand, the solution space of structures is not continuous. The non-differentiability of discrete structured data makes it difficult to perform the optimization by back-propagation, which has shown success in handling continuous data (such as images~\citep{pan2020geometrical}). On the other hand, the search space is combinatorially gigantic, for which we cannot explore exhaustively in a reasonable timeframe~\citep{korte2011combinatorial}.

In early years, the optimization on sequences or graphs is typically accomplished by heuristic search algorithms~\cite{chen2009novel, hwang1988simulated, dorigo1997ant}, based on which some local optima can be captured. Recently, the development of deep neural networks enables researchers to learn a continuous space via generative models and search the desired structures by gradient-based optimization. Gomez-Bombarelli et al.~\cite{gomez2018automatic} use a variational autoencoder (VAE) to encode the molecule sequences and leverage Bayesian optimization to search molecules with desired property profiles. Jin et al.~\cite{junctiontree} introduce a junction tree variational autoencoder for graph generation and adopt a similar continuous optimization strategy. However, as the latent space is high dimensional and the objective functions defined on the latent space are typically non-convex, it is still a difficult task to learn a reliable mapping from the latent space to the property.

Another strategy is to use reinforcement learning (RL) to perform discrete optimization on sequences and graphs. For example, MolDQN~\cite{zhou2019optimization} learns to generate an action sequence that corresponds to a series of modifications on an existing graph structure. Similarly, GCPN~\cite{gcpn} imposes an adversarial loss into the reward system and performs the molecule graph optimization through policy gradient. Considering the rewards are usually obtained based on the property profiles of the final generated molecules, the sparse rewards (per action)  and the large search space hinder the RL agent from figuring out the effective modification action sequences.

\begin{figure}[!t]
	\centering
	\includegraphics[width=0.7\linewidth]{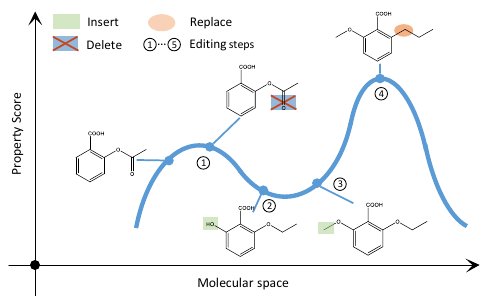}
	\caption{SAGS generates a new molecule by a series of editing operations (i.e., insertion, replacement, and deletion). 
		At each step, SAGS proposes a candidate modification of the sample, which is accepted or rejected according to a certain acceptance rate (only accepted modifications are shown). Although samples are discrete, we make an analogue in the continuous real $x$-axis where the distance of two molecules is roughly given by the number of edits.
	}
	\label{fig:SAGS}
\end{figure}

In this paper, we propose a novel approach that uses  \textit{Simulated Annealing to optimize Graphs and Sequences (SAGS).} The key idea is to integrate powerful neural networks into metaheuristics (e.g., simulated annealing, SA) to restrict the search space in discrete optimization. In our work, we first design a sophisticated objective function, considering both the property profiles of the samples and some necessary constraints (such as grammar validity). Then, we regard the locally edited structures (i.e., via replacement, insertion, and deletion operation) as the analogue of the neighborhoods of a given structure, constituting the sampling candidates of SA. Finally, SAGS searches towards this objective by performing a sequence of local editing steps, where deep generative models propose the editing content. At each editing step, a better candidate (higher scored in the objective) is always accepted by SAGS, while a worse candidate is likely to be rejected, but could also be accepted (controlled by an annealing temperature). At the beginning, the temperature of SA is usually high, and worse samples are more likely to be accepted. This pushes the SA algorithm outside a local optimum. The temperature is cooled down as the optimization proceeds, making the model better settle down to an optimum. Figure~\ref{fig:SAGS} illustrates how SAGS searches for an optimum in molecular optimization. 

We evaluate the effectiveness of our model on five benchmark datasets across text generation and molecular optimization. Experimental results show that SAGS advances the state-of-the-art performance on all these tasks in both automatic metrics and human evaluation. Further investigation on a molecular optimization dataset shows that results provided by the SAGS framework are comparable to the annotated labels. \\

In summary, our contributions are as follows:
\begin{compactitem}
	
	\item  We provides a new perspective to integrate powerful neural networks into metaheuristics (e.g., SA) to enhance the search effectiveness for discrete structures.
	\item  We design a sequence-based SAGS for unsupervised paraphrase generation. We propose an objective function specifically for paraphrasing. We also present a copy mechanism to address the rare words during the generation.
	\item  We design a graph-based SAGS for molecule generation. We propose three primitive editing operations on graphs. We also introduce a sophisticated objective function that allows us to impose both hard and soft constraints. 
	\item We achieve the state-of-the-art performance on four benchmark datasets compared with previous unsupervised paraphrase generators and significantly outperforms previous best molecular generators on various test settings. Ablation studies further confirm the superiority of the SA algorithm in our framework over hill climbing and Metropolis-Hasting.
\end{compactitem}

\section{Related Work}
The optimization of discrete structures has been widely studied for decades~\cite{parker2014discrete}, aiming at generating a new structure with the better property given an existing one. It relates to various domains such as natural language processing (NLP)~\cite{li2020unsupervised}, gene analysis~\cite{hu2018rationalizing}, and drug discovery~\cite{xiao2013optimization}. 

A na\"ive approach to discrete optimization is by hill climbing, which is in fact a greedy algorithm. In NLP, beam search (BS)~\cite{beam} is widely applied to sequence generation. BS maintains a $k$-best list in a partially greedy fashion during left-to-right (or right-to-left) decoding. Further, making distributed modifications (such as the genetic algorithm and Monte-Carlo sampling) over the entire sample has a higher probability of finding the optima. Due to the complexity of the real-world problem (e.g., text generation), such heuristic search algorithms can hardly achieve compelling performances independently.

In recent years, the development of neural networks provides another way to perform optimization on structured data. One of the  mainstream research is to use continuous optimization on a learned representation space of the structured data. For example, Gomez-Bombarelli et al.~\cite{gomez2018automatic} use a variational autoencoder (VAE) to encode the molecule sequences (SMILES strings) and leverage Bayesian optimization to search molecules with desired property profiles. As the sequence generator tends to output invalid SMILES, Kusner et al.~\cite{kusner2017grammar} impose grammar rules into the decoder of VAE and thus improve the quality of generation. However, as the latent space is often high dimensional and the objective functions defined on the latent space are usually non-convex, an agent can hardly learn a reliable mapping from the latent space to the property, limiting the optimization performance of these methods. 

Besides the continuous optimization on the latent space, deep reinforcement learning (RL) learns to generate the structured data in a trial-and-error manner, also serving as a promising approach~\cite{zhou2019optimization,gcpn, Liu-ccgf}.  You et al.~\cite{gcpn} propose to optimize a molecule graph by sequentially modifying a node or an edge. Liu et al.~\cite{Liu-ccgf} model the sequence optimization task in a chance-constrained fashion and use RL to maximize the user-defined rewards. GraphAF~\cite{2020graphaf} introduces a flow-based RL that generates the nodes and edges based on existing sub-graphs. 

Different from these studies, SAGS proposes an optimization framework using edit-based simulated annealing sampling. SAGS further integrates deep generative models to restrict the search space and improve the search effectiveness.


SAGS is also related to editing-based approaches. Miao et al.~\cite{miao2018cgmh} propose to edit a word in a sentence sequentially by Metropolis-Hastings sampling. Schumann et al.~\cite{summarization} search the summarization of a given sentence by introducing a swapping operator on the selected words. Kumar et al.~\cite{kumar2020iterative} generate simplified candidate sentences by iteratively editing the given complex sentence using three simplification operations (i.e., lexical simplification, phrase extraction, deletion and reordering). Guu et al.~\cite{guu2018generating} use a heuristic delete-retrieve-generate approach as a component of a supervised sequence-to-sequence (Seq2Seq) model, but our SAGS is a mathematically inspired, unsupervised search algorithm. Dong et al.~\cite{dong2019editnts} learn the deletion and insertion operations for text simplification in a supervised way, where their groundtruth operations are obtained by some dynamic programming algorithm. Our editing operations (applicable on sequences and graphs) are the search actions during unsupervised simulated annealing.

\red{
In particular, sequence optimization can also be modeled as the sequence-to-sequence (Seq2Seq) problem (e.g., text summarization~\cite{summarization}, paraphrase generation~\cite{gupta2018deep} and sequence labeling~\cite{shao2021self}). Following this direction, several effective models, including stacked residual LSTM~\cite{gupta2018deep} and the Transformer model~\cite{wang2019a} were introduced. However, these models usually demand enormous parallel data. By contrast, our method is unsupervised and can be flexibly applied to different domains.
}
\section{Approach}

In this section, we present our novel SAGS framework that uses simulated annealing (SA) for the optimization of sequences and graphs. In particular, we first present the general SAGS algorithm, and then design a sequence-based SAGS for unsupervised paraphrasing and a graph-based SAGS for molecule generation separately.

\subsection{The General SAGS Framework}

We introduce our general Simulated Annealing framework that can work for both Graph and Sequence optimization (SAGS). 

Simulated Annealing (SA) is an effective and general metaheuristic of search, especially for a large discrete or continuous space~\cite{Kirkpatrick671}. At each time step, SA randomly selects a solution and accepts it according to a probability that considers its quality and the current temperature. During the search, the probability remains at one for a better solution and gradually decreases to zeros for a worse solution. Theoretically, simulated annealing is guaranteed to converge to the global optimum in a finite problem if the proposal and the temperature satisfy some mild conditions~\cite{granville1994simulated}. Although such convergence may be slower than exhaustive search and the sample space is, in fact, potentially infinite, simulated annealing is still a widely applied search algorithm~\cite{fleischer1995simulated,fayyaz2018simulated}, especially for discrete optimization. Readers may refer to Hwang~\cite{hwang1988simulated} for details of the SA algorithm. 

Generally speaking, SAGS searches the discrete structure space towards a pre-defined objective by performing a series of local edits via SA (Algorithm~\ref{alg:SAGS}). Specifically, let $\mathcal{X}$ be a (huge) search space of a kind of discrete and structured data, such as graphs, and $f(\mathrm x)$ be an objective function. The goal is to search for a sample $\mathrm x$ that maximizes $f(\mathrm x)$. At a search step $t$, SAGS keeps a current sample $\mathrm x_t$, and proposes a new candidate $\mathrm x_*$ by the candidate sample generator via the three primitive editing operations (i.e., insertion, replacement and deletion). If the new candidate is better scored by $f$, i.e., $f(\mathrm x_*)>f(\mathrm x_t)$, then SAGS accepts the proposal. Otherwise, SAGS tends to reject the proposal $x_*$, but may still accept it with a small probability $e^{\frac{f(\mathrm x_*)-f(\mathrm x_t)}{T}}$, controlled by an annealing temperature $T$. In other words, the probability of accepting the proposal is
\begin{align}
\label{eq:temp}
    p(\text{accept}|\mathrm x_*,\mathrm x_t, T) = \min\big(1, e^{\frac{f(\mathrm x_{*})-f(\mathrm x_{t})}{T}}\big).
\end{align}
If the proposal is accepted, $\mathrm x_{t+1}=\mathrm x_*$, or otherwise, $\mathrm x_{t+1}=\mathrm x_t$.

Inspired by the annealing in chemistry, the temperature $T$ is usually high at the beginning of search, leading to a high acceptance probability even if $\mathrm x_*$ is worse than $\mathrm x_t$.  Then, the temperature is decreased gradually as the search proceeds. In our work, we adopt the linear annealing schedule, given by $T = \max(0, T_\text{init}- C\cdot t)$, where  $T_\text{init}$ is the initial temperature and $C$ is the decreasing rate. The high initial temperature of SA makes the algorithm less greedy compared with hill climbing, whereas the decreasing of temperature along the search process enables it to better settle down to a certain optimum.

\begin{algorithm}[t]
	\caption{The SAGS framework}\label{alg:SAGS}\footnotesize
	
	\begin{algorithmic}
		\State \!\!\!\!1:\ \textbf{Input}: Original sample $\mathrm x_0$
		\State \!\!\!\!2:\ \textbf{for} $t\in \{1,\dots,N\}$ \textbf{do}
		\State \!\!\!\!3:\ \ \ \ $T =\max\{T_{\text{init}}- C\cdot t,0\}$
		\State \!\!\!\!4:\ \ \ \ Candidate sample generator randomly chooses an editing operation (denoted by op) and a position $k$
		\State \!\!\!\!5:\ \ \ \ Candidate sample generator edits the current sample $x_t$ at position $k$
		\State \!\!\!\!6:\ \ \ \  Generative models rank the modified samples and select $M$ samples that most likely exist in nature
		\State \!\!\!\!7:\ \ \ \  Candidate sample generator samples a candidate $\mathrm x_*$ from the $M$ samples
	\State \!\!\!\!8:\ \ \ \ Compute the accepting probability $p_{\text{accept}}$ by Eqn.~(\ref{eq:temp})
		\State \!\!\!\!9:\ \ \ \ With probability $p_{\text{accept}}$,  $\mathrm  x_{t+1} = \mathrm x_*$
		\State \!\!\!\!\!\!\!10:\ \ \ \ With probability $1-p_{\text{accept}}$,  $\mathrm x_{t+1} = \mathrm x_t$
		\State \!\!\!\!\!\!\!11:\ \textbf{end for}
		\State \!\!\!\!\!\!\!12:\ \textbf{return} $\mathrm x_\tau$ s.t. $\tau=\text{argmax}_{\tau\in \{1,\dots,N\}} f(\mathrm x_\tau)$
\end{algorithmic}

\end{algorithm}

\textbf{Alternative Metaheuristics.} As introduced previously, the key idea of our work is to integrate powerful neural networks into metaheuristics to restrict the search space in graph and sequence optimization. Simulated annealing is just a representative example of them. Users can also apply other heuristic algorithm into our framework, such as hill climbing and genetic algorithm. Some of swarm intelligence algorithms are not compatible with our framework, for example, particle swarm optimization (PSO)~\cite{kennedy1995particle} and ant colony optimization (ACO)~\cite{dorigo1997ant}, because they works on the continuous space. Different from the above algorithms, SA is a flexible optimization algorithm that just posits the sampling process happens on neighbors rather than specifying concrete sampling operations (such as crossover in GA). Thus, SA allows us to contribute a more suitable design (e.g., edit operations) according to a given task. In addition, SA is mathematically inspired and is guaranteed to converge to the global optimum when some mild conditions are satisfied. 

As individual discrete structures have different optimization goals and editing operations, below, we will elaborate two versions of SAGS for the sequence and graph optimization, respectively. 

\subsection{SAGS for Unsupervised Paraphrasing as Sequence Optimization}
We take unsupervised paraphrase generation as an example to elucidate our SAGS framework for the sequence optimization task.  Paraphrasing aims to restate one sentence as another with the same meaning, but different wordings. It constitutes a corner stone in many NLP tasks,  such as question answering, information retrieval, and dialogue systems. 
However, automatically generating accurate and
different-appearing paraphrases is a still challenging research problem, due to the complexity of natural language. 

Conventional approaches~\cite{prakash2016neural,gupta2018deep} model the paraphrase generation as a supervised encoding-decoding problem, inspired by machine translation systems. Usually, such models require massive parallel samples for training. In machine translation, for example, the WMT 2014 English-German dataset contains 4.5M sentence pairs. However, the training corpora for paraphrasing are usually small and domain-specific. The widely-used Quora dataset\footnote{https://www.kaggle.com/c/quora-question-pairs} only contains question sentences with a size of 140K pairs. Thus, supervised paraphrase models do not generalize well to new domains~\cite{zichao2019}.

As a result, unsupervised methods would largely benefit the paraphrase generation task since no parallel data are needed. With the help of deep learning, researchers are able to generate paraphrases by sampling from a neural network-defined probabilistic distribution, either in a continuous latent space~\cite{bowman2015generating} or directly in the word space~\cite{miao2018cgmh}. However, the meaning preservation and expression diversity of those generated paraphrases are less ``controllable'' in such probabilistic sampling procedures.

To this end, we apply SAGS to the unsupervised paraphrasing task. In our work, we design a sophisticated objective function, considering semantic preservation, expression diversity, and language fluency of paraphrases. SAGS searches towards this objective by performing a sequence of local editing steps, namely, word replacement, insertion, and deletion. For each step, the candidate sample generator in SAGS first proposes a potential editing, and SAGS then accepts or rejects the proposal based on sample quality (as introduced previously). In the following, we will further elaborate the two primary components in SAGS for unsupervised paraphrasing, that is, the objective function and the candidate sample generator.

\subsubsection{Objective Function}
Simulated annealing maximizes an objective function, which can be flexibly specified in different applications.
In particular, our objective $f(\mathrm x)$ for unsupervised paraphrasing considers multiple aspects of a candidate paraphrase, including semantic preservation, expression diversity, and language fluency. Thus, our search objective is to maximize
\begin{align}
\label{eq:obj}
 f(\mathrm x) &= f_{\text{sem}}(\mathrm x,\mathrm x_0) \cdot f_{\text{exp}}(\mathrm x,\mathrm x_0)\cdot f_{\text{flu}}(\mathrm x),
\end{align}
where $\mathrm x_0$ is the input sentence. In the objective function, the semantic preservation function $f_\text{sem}$ evaluates the semantic similarity between the candidate paraphrase $\mathrm x$ and the input sentence $\mathrm x_0$. The expression diversity function $f_{\text{exp}}$ calculates the expression difference between the candidate and the input sentence. The language fluency function $f_{\text{flu}}$ computes the probability of the candidate paraphrase being a natural sentence. The implementations of these objective functions are described below.

\textbf{Semantic Preservation.} A paraphrase is expected to capture all the key semantics of the original sentence. Thus, we leverage the cosine function of keyword embeddings to measure if the key focus of the candidate paraphrase is the same as the input. Specifically, we extract the keywords of the input sentence $\mathrm x_0$ by the Rake system~\citep{rose2010automatic} and embed them by GloVE~\citep{pennington2014glove}. For each keyword, we find the closest word in the candidate paraphrase $\mathrm x_*$ in terms of the cosine similarity. Our keyword-based semantic preservation score is given by the lowest cosine similarity among all the keywords, i.e., the least matched keyword:
 \begin{align}
    f_{\text{sem,key}}(\mathrm x_{*},\mathrm x_0)=\!\!\!\min_{ e\in \text{keywords}(\mathrm x_0)}\!\!\!\max_j\{\cos(\bm w_{*,j},\bm e)\},
 \end{align}
where  $w_{*,j}$ is  the  $j$th word in the sentence $\mathrm x_*$; $e$ is an extracted keyword of $\mathrm x_0$. Bold letters indicate embedding vectors.

In addition to keyword embeddings, we also adopt a sentence-level similarity function, based on Sent2Vec embeddings~\citep{pagliardini2017unsupervised}.
Sent2Vec learns $n$-gram embeddings and computes the average of $n$-grams embeddings as the sentence vector. It has been shown to be significant improvements over other unsupervised sentence embedding methods in similarity evaluation tasks~\citep{pagliardini2017unsupervised}. 
Let ${\mathbf x}_*$ and  $\mathbf x_0$ be the Sent2Vec embeddings of the candidate paraphrase and the input sentence, respectively. Our sentence-based semantic preservation scoring function is $f_{\text{sim,sen}}(\mathrm x_{*},\mathrm x_0)=\cos({\mathbf x}_{*},{\mathbf x}_0)$.

To sum up, the overall semantic preservation scoring function is given by
\begin{align}
    f_{\text{sem}}(\mathrm x_{*},\mathrm x_0)=  f_{\text{sem,key}}(\mathrm x_{*},\mathrm x_0)^P\cdot f_{\text{sem,sen}}(\mathrm x_{*},\mathrm x_0)^Q,
 \end{align}
where $P$ and $Q$ are hyperparameters, balancing the importance of the two factors. Here, we use power weights because the scoring functions are multiplicative.  

\textbf{Expression Diversity.}
The expression diversity scoring function computes the lexical difference of two sentences. We adopt a BLEU-induced function to penalize the repetition of the words and phrases in the input sentence:
\begin{align}
   f_{\text{exp}}(\mathrm x_{*},\mathrm x_0)=(1-\text{BLEU}(\mathrm x_*,\mathrm x_0))^S,
\end{align}
where the BLEU score computes a length-penalized geometric mean of $n$-gram precision ($n=1,\cdots,4$). $S$ coordinates the importance of $f_{\text{exp}}(\mathrm x_{t},\mathrm x_0)$ in the objective function (\ref{eq:obj}).

\textbf{Language Fluency.} Despite semantic preservation and expression diversity, the candidate paraphrase should be a fluent sentence by itself. We use a separately trained (forward) language model (denoted as $\overrightarrow{\text{LM}}$) to compute the likelihood of the candidate paraphrase as our fluency scoring function:
\begin{align}
   f_{\text{flu}}(\mathrm x_{*})=\prod_{k=1}^{k=l_*} p_{\overrightarrow{\text{LM}}}( w_{*,k}|  w_{*,1},\dots, w_{*,k-1}),
\end{align}
where $l_*$ is the length of $\mathrm x_*$ and $w_{*,1},\dots, w_{*,l}$ are words of $\mathrm x_*$.
Here, we use a dataset-specific language model, trained on  non-parallel sentences. Notice that a weighting hyperparameter is not needed for  $f_{\text{flu}}$, because the relative weights of different factors in Eqn.~(\ref{eq:obj}) are given by the powers in $f_\text{sem,key}$, $f_\text{sem,sen}$, and $f_\text{exp}$.

\subsubsection{Candidate Sample Generator for Sequences}
\label{sec:generation}
At each search action, SAGS proposes a candidate sample by a local modification, which is either accepted or rejected by Equation~(\ref{eq:temp}). Since each action yields a new sample $\mathrm x_*$ from $\mathrm x_t$, we call it a \textit{candidate sample generator}. While the proposal of candidate samples does not affect convergence in theory (if some mild conditions are satisfied), it may largely influence the efficiency of SA search.

To make a proposal, SAGS provides three primitive operators on the structure of the given sample $\mathrm x_t$, namely the replacement, insertion and deletion. Let the current sequence sample be $\mathrm x_t=(w_{t,1},\dots, w_{t,k-1},$ $w_k,w_{t,k+1}\dots, w_{t,l_t})$. If the replacement operation proposes a candidate word $w_*$ for the $k$th element, the resulting candidate sample becomes $\mathrm x_*=(w_{t,1},\dots, w_{t,k-1},$ $ w_*,w_{t,k+1}\dots, w_{t,l_t})$, where $l_t$ stands for the length of sequence sample. The insertion operation and deletion one work similarly.

Here, the edited sample $x_*$ is sampled from a probabilistic distribution, induced by the objective function (\ref{eq:obj}):
 \begin{align}
     p(x_*|\cdot)&=
\frac{f(x_*)}{\sum_{w_{\dag}\in\mathcal{W}}f(x_\dag)},
\label{eq:prob_func}
 \end{align}
where $\mathcal{W}$ is the sampling vocabulary. We observe that sampling from such objective-induced distribution typically yields a meaningful candidate sample, which enables SA to explore the search space more efficiently.

It is also noted that, when the sampling vocabulary is gigantic (such as in the natural language processing), sampling an element from the entire vocabulary involves re-evaluating function~(\ref{eq:obj}) for each candidate element. Therefore, we follow Miao et al.~\cite{miao2018cgmh} and only sample from the top-$K$ elements given by jointly considering a forward generative model and backward generative model (i.e., the language model). The replacement operator, for example, suggests the vocabulary of the top-$K$ elements by 
\begin{align}
    \mathcal{W}_{t,\text{replace}}= \operatorname{top-}K_{w_*} \Big[p_{\overrightarrow{\text {LM}}}(w_{t,1},\dots,w_{t,k-1},w_*)  \cdot  p_{\overleftarrow{\text {LM}}}(w_*,w_{t,k+1},\dots,w_{t,l_t})\Big].
\end{align}
For the insertion operator, the top-$K$ vocabulary $\mathcal{W}_{t,\text{insert}}$ is computed in a similar way (except that the position of $w_*$ is slightly different). Details are not repeated. 

\textbf{Copy Mechanism.}
When the element vocabulary is large, we also observe that rare elements are sometimes deleted or replaced during SA stochastic sampling. They may be important but are difficult to be recovered because they usually have a low language model-suggested probability. 

Therefore, we propose a copy mechanism for SA sampling. Specifically, we allow the candidate sample generator to copy the elements from the original sample $\mathrm x_0$ for replacement and insertion operators. This is essentially enlarging the top-$K$ sampling vocabulary with the words in $\mathrm x_0$, given by
\small
\begin{align}
\widetilde{\mathcal{W}}_{t,\text{op}} = \mathcal{W}_{t,\text{op}} \cup \{ w_{0,1},\dots,  w_{0,l_0}\};\ \ 
\text{op}\! \in\! \{\text{replace,insert}\}
\end{align}
\normalsize
{Thus, $\widetilde{\mathcal{W}}_{t,\text{op}}$ is the actual vocabulary from which SA samples the word $w_*$ for replacement and insertion operation. }

While such vocabulary reduces the proposal space, it works well empirically because other low-ranked candidate words are either irrelevant or makes the sample far from the data distribution (measured by the the language model); they usually have low objective scores, and are likely to be rejected even if sampled. 

\subsection{SA for Molecule Generation as Graph Optimization}
We take molecule generation as an example to elucidate our SAGS framework for graph optimization. Molecule generation, as one of the directions for drug discovery, is an important task in the field of bioinformatics. Here, we consider a specific molecule optimization application, which aims to generate a new molecule that is similar to a given one but has better hydrophobicity. The hydrophobicity measures how likely a molecule can mix with the organic compound, which is an important factor in drug interactions with biological membranes. Considering that constructing a parallel dataset for this task is expensive and labor-intensive, we also focus on unsupervised molecule generation.

Existing methods for unsupervised molecule generation can be grouped into two categories: 1) gradient-based methods that optimize the molecules based on a learned space~\cite{cvae} and 2) reinforcement learning methods (e.g., GraphAF~\cite{2020graphaf} and CGPN~\cite{you2018graph}) that learn to generate the desired molecules in an auto-regressive manner. But the methods in the first category, such as CVAE~\cite{cvae} and GVAE~\cite{kusner2017grammar}, tend to generate invalid molecules.  On the other hand, the RL methods are still challenged by the sparse reward per action during the generation of the molecule graph. 

Therefore, an effective graph optimization method is an urgent need for molecule generation. In this work, we propose a graph-based SAGS for molecule generation. Similar to the sequence-based SAGS for unsupervised paraphrasing, the graph-based SAGS also follows the general framework and has specific implementations in the objective function and the candidate sample generator, both of which will be described in detail below.

\subsubsection{Objective Function}
According to the goal of molecule generation, we need to maximize both the property score and the similarity of the newly generated molecule $x_*$, which are reflected in a function $f_p$, given by
\begin{align}
f_p(\mathrm x_*,x_0) &= f_\text{LogP}(x_*)+Mf_\text{sim}(x_*,x_0),
\end{align}
where $f_\text{LogP}(x_*)$ indicates the penalized logP score of the molecule $x_*$. As in Shi et al.~\cite{you2018graph}, we measure the hydrophobicity of the molecule via the logP score penalized by the ring size and synthetic accessibility. The penalized logP score (also denoted by PlogP) can be accurately approximated by a machine learning predictor that takes a number of atomic features (such as the type of aromatic group) as input~\cite{wildman1999prediction,li2021trimnet}. $f_\text{sim}$ measures the Morgan similarity~\cite{roger} of the new molecule to the input molecule, and $M$ is the hyper-parameter.

At the same time, the generation of SAGS should conform to the constrains including the similarity condition and valency conditions of the molecule. Therefore, we use another function to represent the overall validity of the generated molecule, given by
\begin{align}
	\text{Validity}(\mathrm x_*,x_0) &= f_g(x_*)\cdot \textbf{1}_{\{f_\text{sim}(x_*,x_0)>\delta\}},
\end{align}
where the function $f_g$ indicates  whether the sample $x_*$ is subject to the valency conditions of the molecules or not, which can be directly calculated by RDKit package. where $\textbf{1}_{\{\cdot\}}$ is an indicator function that yields 1 when its argument is true, and 0 otherwise. In other words, a newly generated molecule $x_*$ is regarded as valid only if its grammar is correct and the similarity to the original molecule is larger than $\delta$.

Finally, we combine these two functions (i.e., $f_p$ and the validity function) and define the objective of SAGS for molecule generation as a piecewise function, given by
\begin{align}
f(x_*)&=\left\{
\begin{aligned}
 & f_p(\mathrm x_*,x_0),\quad \text{if} \quad \text{Validity}(x_*,x_0)> 0, \\
 & -\infty,\quad \text{if} \quad \text{Validity}(x_*,x_0)\le 0,
\end{aligned}
\right.
\label{eq:graphobj}
\end{align}
where we prevent the invalid molecules to be sampled by assigning negative infinity to their objective values.

\subsubsection{Candidate Sample Generator for Graphs}
\begin{figure}[!hbt]
	\centering
	\includegraphics[width=0.9\linewidth]{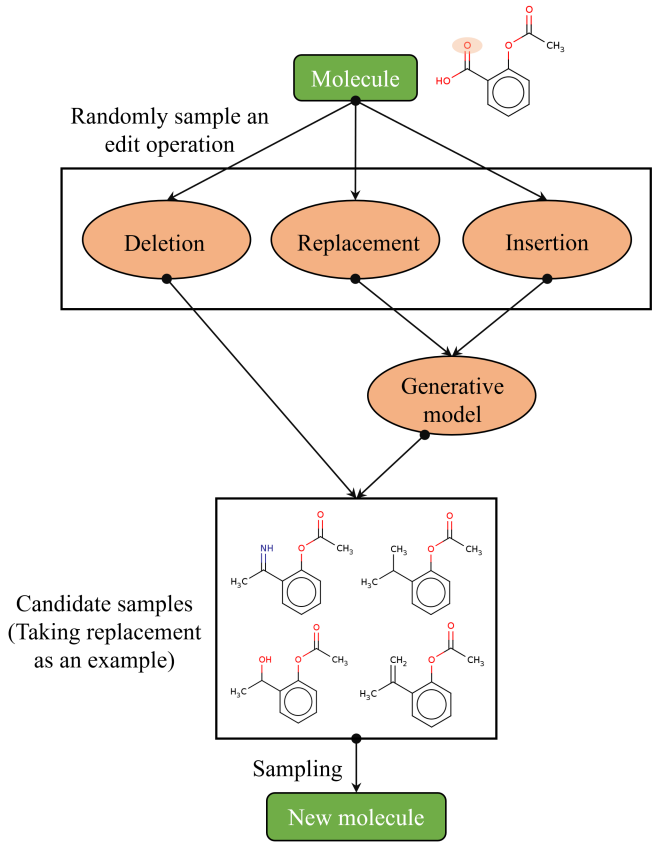}
	\caption{The graph editing process of a particular iteration step.}
	\label{fig:graph-design}
\end{figure}

Let the current graph sample be $x_t=\{V_t,E_t\}$, where $V_t=(v_{t,1},\cdots,v_{t,k},\cdots,v_{t,n})$ and $E_t$ is a set of nodes and edges, $n$ represents the number of nodes. As shown in Figure~\ref{fig:graph-design}, the candidate sample generator yields a new graph by modifying the current graph sample. The modification is based on a randomly sampled position $k$. If the replacement operation proposes a candidate node $v_*$, the resulting candidate sample becomes $x_*=\{V_*=(v_{t,1},\cdots,v_{t,k-1},v_{*},\cdots,v_{t,n}),E_t\}$. If the insertion operation proposes a candidate node $v_*$, the nodes of the graph become $V_*=(v_{t,1},\cdots,v_{t,k},v_{*},\cdots,v_{t,n})$; Meanwhile, the insertion operation also adds a new edge $e_{(k,*)}$ to connect the newly formed $v_*$ with $v_{t,k}$, and thus the resulting candidate sample become $x_*=\{V_*,E_t\cup e_{(k,*)}\}$. The deletion operation for graphs works similarly. In the particular cases where the atom to be deleted lies on the ring, the deletion operation will break the ring. Meanwhile, we also include a new ring where its two neighboring atoms are connected. That is to say, the deletion operation will result in two candidate molecules, one molecule with the ring broken and the other with a smaller ring.

Similar to candidate sample generator for sequences, the new graph $x_*$ is sampled from a probabilistic distribution by Equation (\ref{eq:soft_func}). 
\begin{align}
     p(x_*|\cdot)&=
\frac{e^{f(x_*)}}{\sum_{w_{\dag}\in\mathcal{W}}e^{f(x_\dag)}},
\label{eq:soft_func}
 \end{align}
where $\mathcal{W}$ is the sampling vocabulary of candidate nodes. Note that we use softmax function to produce the sampling distribution because the objective values can be negative in some cases.

To accelerate the inference speed, we restrict the vocabulary $\mathcal{W}$ to a set of the top-K nodes given by a generative graph neural network. For example, the replacement operator suggests the top-K nodes by
\begin{align}
    \mathcal{W}_{t,\text{replace}}= \operatorname{top-}K_{v_*}p(\{V_t\backslash v_{t,k},E_t\},v_*)
    \label{eq:graph_node}
\end{align}
where $V_t\backslash v_{t,k}$ represents the node set which removes the node $v_{t,k}$. The function $p(\{V_t\backslash v_{t,k},E_t\},v_*)$ stands for the likelihood of the modified graph approximated by a graph generative model. We adopt the molecular pre-training graph neural network (MPG)~\cite{libbab109} as the generative model. MPG pre-trains a graph neural network called MolGNet on the ZINC250K dataset with a masking strategy to provide a probabilistic distribution for the masked nodes, serving as one of the most advanced architecture for graph structure modeling~\cite{libbab109}. The sampling vocabulary of the candidate nodes for insertion operator is computed in a similar way.

\section{Experiments on Paraphrase Generation}
\label{s:dataset}
\subsection{Datasets}  

   \textbf{Quora.} The Quora question pair dataset contains 140K parallel paraphrases and additional 260K pairs of non-parallel samples. We follow the unsupervised setting in Miao et al.~\cite{miao2018cgmh}, where 3K and 20K pairs are used for validation and test, respectively. 
   
   	 \textbf{Wikianswers.} The original Wikianswers dataset~\citep{fader2013paraphrase} contains 2.3M pairs of question paraphrases from the Wikianswers website. Since our model only involves training a language model, we randomly selected 500K non-parallel samples for training. For evaluation, we followed the same protocol as Li et al.~\cite{zichao2019} and randomly sampled 5K for validation and 20K for testing. Although the exact data split in previous work is not available, our results are comparable to previous ones in the statistical sense. 
   	 
  \textbf{MSCOCO.} The MSCOCO dataset contains 500K+ paraphrases pairs for $\sim$120K image captions~\citep{lin2014microsoft}. We follow the standard split and the evaluation protocol in Prakash et al.~\cite{prakash2016neural} where only image captions with fewer than 15 words are considered, since some captions are extremely long (e.g., 60 words).
  
  \textbf{Twitter.} The Twitter URL paraphrasing corpus~\citep{lan2017a} is originally constructed for paraphrase identification. We follow the standard train/test split, but take 10\% of the training data as the validation set. The remaining samples are used to train our language model. For the test set, we only consider sample pairs that are labeled as ``paraphrases.'' This results in 566 test cases. 
  
\subsection{Competing Methods} 
In paraphrase generation, we would compare SAGS with recent discrete and continuous sampling-based paraphrase generators, namely, VAE, Lag VAE~\citep{he2018lagging}, and CGMH. Early work on unsupervised paraphrasing typically adopts rule-based methods~\citep{Barzilay-rules}. Their performance could not be verified on the above datasets, since the extracted rules are not available. Therefore, we are unable to compare them in this paper. Also, rule-based systems usually do not generalize well to different domains. In the following, we describe our competing methods:

\textbf{VAE.} We train a variational autoencoder (VAE) with two-layer, 300-dimensional LSTM units\footnote{We used the code in https://github.com/timbmg/sample-VAE}. The VAE is trained with non-parallel corpora by maximizing the variational lower bound of log-likelihood; during inference, samples are sampled from the learned variational latent space~\citep{bowman2015generating}.

\textbf{Lag VAE.} He et al.~\cite{he2018lagging} propose to aggressively optimize the inference process of VAE with more updates to address the posterior collapse problem. This method has been reported to be the state-of-the-art VAE. We adopted the published source code and generated paraphrases for comparison.

\textbf{CGMH.} Miao et al.~\cite{miao2018cgmh} use Metropolis--Hastings sampling in the word space for constrained sample generation. It is shown to outperform latent space sampling as in VAE, and is the state-of-the-art unsupervised paraphrasing approach. We also adopted the published source code and generated paraphrases for comparison.

\textbf{SAGS (GA).} We provide a GA (genetic algorithm) version of our framework to investigate the different choices of metaheuristics in our framework. Different from the traditional GA, SAGS (GA) adopts the generative neural network to implement the mutation operation. As for the crossover operation in SAGS (GA), we first randomly select a position (i.e., a word in sentences) to split a sample into two fragments and then combine the fragments from different samples to constitute a new sample in the population. To have a fair comparison, SAGS (GA) shares the same objective function and generative model with SAGS. The hyperparameters of SAGS (GA) involve the population size, the probability of mutation, the probability of crossover and the iteration number. All the above hyperparameters are calibrated by the grid search scheme which are similar to the one of SAGS.

We further compare SAGS with supervised Seq2Seq paraphrase generators: ResidualLSTM~\citep{prakash2016neural}, VAE-SVG-eq~\citep{gupta2018deep}, Pointer-generator~\citep{see2017get}, the Transformer~\citep{vaswani2017attention}, and the decomposable neural paraphrase generator (DNPG)~\cite{zichao2019}. DNPG has been reported as the state-of-the-art supervised paraphrase generator. To better compare SAGS with all paraphrasing settings, we also include domain-adapted supervised paraphrase generators that are trained in a source domain but tested in a target domain, including shallow fusion~\citep{gulcehre2015using} and
multi-task learning (MTL)~\cite{domhan2017using}.

\subsection{Metrics} 
For paraphrase generation, we adopt BLEU~\citep{papineni2002bleu} and ROUGE~\citep{lin2004rouge} scores as automatic metrics to evaluate model performance. Sun and Zhou~\cite{sun2012joint} observe that BLEU and ROUGE could not measure the diversity between the generated and the original samples, and propose the iBLEU variant by penalizing by the similarity with the original sample. Therefore, we regard the iBLEU score as our major metric, which is also adopted in Li et al.~\cite{zichao2019}. In addition, we also conduct human evaluation in our experiments (detailed later).

\subsection{Implementation Details}  

Our method for paraphrasing involves unsupervised language modeling (forward and backward), realized by two-layer LSTM with 300 hidden units and trained specifically on each dataset with non-parallel samples. Besides, SAGS involves the power weights  $P, Q,$ and $S$ of the objective function, the initial temperature $T_\text{init}$ and the cooling coefficient $C$. The values of the power weights $P, Q,$ and $S$ coordinate the optimization direction of SAGS. We should find a best combination that guides the search process towards a reasonable paraphrase. The initial temperature $T_\text{init}$ and the cooling coefficient $C$ jointly control the search range and search speed of SAGS. We also need to calibrate them to make SAGS as efficient as possible. Therefore, we followed the conventional hyperparameter tuning procedure in machine learning~\cite{zhang2017sensitivity,li2017paraphrase}, i.e., the grid search, to determine the values of these hyperparameters. 

The grid search procedure was performed on the validation set of the Quora dataset using the iBLEU metric. In particular, considering the scoring functions in the objective are multiplicative, the range of the power weights $P, Q,$ and $S$ is set to conform a geometric form, that is, $\{0.5,1,2,4,8\}$. Besides, the initial temperature $T_\text{init}$ was chosen from $\{0.5,1,3,5,7,9\}\times10^{-2}$ and the cooling coefficient $C$ was chosen from  $\{0.5,1,3,5,7,9\}\times10^{-4}$. The grid search procedure aims at finding the best combination of these five hyperparameters. Therefore, we scanned all the combinations of these hyperparameters in terms of the paraphrasing performance. Table~\ref{tab:hyper-t} and Table~\ref{tab:hyper-2} present several representative results of the hyperparameter tuning, where we noticed the power weight of the expression diversity has the highest impact on the paraphrasing performance. We also observed that best combination of these hyperparameters were $P=8, Q=S=1, T_\text{init}=0.03$ and $C=3\times10^{-4}$, which were used in the testing of SAGS.

We should emphasize that all hyper-parameters for paraphrasing were validated only on the Quora dataset, and we did not perform any tuning on the other datasets (except the language model). This shows the robustness of our SAGS model and its hyperparameters.

\begin{table*}[t]	
	\begin{center}
	\resizebox{0.95\linewidth}{!}{
		\begin{tabular}{lllllllllll}
			\hline\noalign{\smallskip}
				   \multirow{2}*{} & & \multicolumn{4}{c}{Quora} &\multicolumn{4}{c}{Wikianswers}\\
		    \cmidrule(r){3-6}\cmidrule(r){7-10}
			
		    & Model & iBLEU & BLEU  & Rouge1 &  Rouge2 & iBLEU & BLEU  & Rouge1 &  Rouge2 \\
			\hline
			\noalign{\smallskip}
			\multirow{6}*{Supervised} 
			& ResidualLSTM      & 12.67 & 17.57 & 59.22 & 32.40 & 22.94 & 27.36  & 48.52 & 18.71\\
			& VAE-SVG-eq        & 15.17 & 20.04 & 59.98 & 33.30 & 26.35  & 32.98 & 50.93 & 19.11\\
			& Pointer-generator & 16.79 & 22.65 & 61.96 & 36.07 & 31.98 & 39.36  & 57.19 & 25.38   \\
    		& Transformer       & 16.25 & 21.73 & 60.25 & 33.45 & 27.70  & 33.01 & 51.85 & 20.70   \\
    		& Transformer+Copy  & 17.98 & 24.77 & 63.34 & 37.31  & 31.43 & 37.88 & 55.88 & 23.37 \\
    		& DNPG               & \underline{\textbf{18.01}} & \underline{\textbf{25.03}}  & \underline{\textbf{63.73}} & \underline{\textbf{37.75}} & \underline{\textbf{34.15}} & \underline{\textbf{41.64}}  & \underline{\textbf{57.32}} & \underline{\textbf{25.88}}  \\
			\noalign{\smallskip}
			\hline
			\noalign{\smallskip}
			\multirow{4}*{Supervised } 
			& Pointer-generator  & 5.04 &  6.96  & 41.89  & 12.77 & 21.87  & 27.94 & 53.99 & 20.85   \\
			& Transformer+Copy   & 6.17 &  8.15  & 44.89  & 14.79  & 23.25 & 29.22  & 53.33 & 21.02\\
			& Shallow fusion     & 6.04 &  7.95 & 44.87  & 14.79   &22.57  & 29.76 & 53.54 &
 20.68 \\
			\multirow{1}*{+ Domain-adapted} 			
			& MTL  & 4.90 &  6.37  & 37.64 & 11.83 & 18.34 & 23.65  & 48.19 & 17.53 \\
			& MTL+Copy  & 7.22 & 9.83  & 47.08 & 19.03 &21.87 & 30.78 & 54.10 & 21.08 \\
			& DNPG  & \underline{10.39} &  \underline{16.98}  & \underline{56.01} & \underline{28.61}  & \underline{25.60} & \underline{35.12} & \underline{56.17} & \underline{23.65} \\
			\noalign{\smallskip}
			\hline
			\noalign{\smallskip}
			\multirow{4}*{Unsupervised } 
			& VAE   & 8.16 &  13.96  & 44.55 & 22.64 & 17.92 & 24.13  & 31.87 & 12.08 \\
			& Lag VAE & 8.73 & 15.52 & 49.20 & 26.07 & 18.38 & 25.08 & 35.65 & 13.21\\
			& CGMH  & 9.94 & 15.73   & 48.73 & 26.12 & 20.05 & 26.45  & 43.31  & 16.53 \\
			& SAGS (GA)  & 6.43 & 9.24   & 28.89 & 15.92 & 15.14 & 18.83  & 27.13  & 11.04 \\
			& SAGS  & \underline{12.03}* & \underline{18.21}*   & \underline{59.51}* & \underline{32.63}* & \underline{24.84}* & \underline{32.39}*   & \underline{54.12}* & \underline{21.45}*  \\
			\noalign{\smallskip}
			\hline
		\end{tabular}
	}
	\end{center}
		\caption{Performance on the Quora and Wikianswers datasets. The best scores within the same training setting are underlined. \red{*: P$<$0.01 by the Friedman and postdoc Nemenyi test in the comparison between SAGS and the other unsupervised baselines, respectively. See Table~\ref{table:test-text} for more statistical results.}}
	\label{table:wiki}
\end{table*}

\begin{table*}[t]
 \begin{center}
	\resizebox{0.8\linewidth}{!}{
		\begin{tabular}{llllllllll}
		\hline
		    \noalign{\smallskip}
		   \multirow{2}*{Model} & \multicolumn{4}{c}{MSCOCO} &\multicolumn{4}{c}{Twitter}\\
		    \cmidrule(r){2-5}\cmidrule(r){6-9}
		&     iBLEU    	&       BLEU     &       Rouge1     &       Rouge2    
		&       iBLEU    	&       BLEU      &       Rouge1     &       Rouge2    \\
			\noalign{\smallskip}
		    \hline
			\noalign{\smallskip}
			VAE & 7.48 & 11.09    & 31.78 & 8.66     & 2.92 & 3.46  & 15.13 & 3.40   \\
			Lag VAE & 7.69 & 11.63    & 32.20 & 8.71     & 3.15 & 3.74  & 17.20 & 3.79   \\
			CGMH & 7.84  & 11.45  & 32.19    & 8.67    & 4.18   & 5.32   & 19.96 & 5.44   \\
			SAGS & \textbf{9.26}* &  \textbf{14.16}*    & \textbf{37.18}* &  \textbf{11.21}*  & \textbf{4.93}* & \textbf{6.87}*   &   \textbf{28.34}* & \textbf{8.53}*\\
			\hline
		\end{tabular}
		}
		\end{center}
	\caption{Performances on MSCOCO and Twitter. \red{*: P$<$0.01 by the Friedman and postdoc Nemenyi test in the comparison between SAGS and the other baselines, respectively.}}
	\label{table:coco}
\end{table*}

\subsection{Results}

Table~\ref{table:wiki} presents the performance of all competing methods on the Quora and Wikianswers datasets. The unsupervised methods are only trained on the non-parallel samples. The supervised models were trained on 100K paraphrase pairs for Quora and 500K pairs for Wikianswers. The domain-adapted supervised methods are trained on one dataset (Quora or Wikianswers), {adapted using non-parallel text on the other (Wikianswers or Quora), and eventually tested on the latter domain (Wikianswers or Quora).}

We observe in Table~\ref{table:wiki} that, among unsupervised approaches, SAGS (GA) obtains the worst scores in terms of all the evaluation metrics, indicating that GA is not a suitable alternative meta-heuristic algorithm for our framework. We also observe that the samples produced by the crossover operation often die out (unselected into the next generation due to low fluency) during the evolution process, which may be the reason for its lackluster performances. VAE and Lag VAE also achieve poor performance on both datasets, indicating that paraphrasing by latent space sampling is worse than word editing. We further observe that SAGS yields significantly better results than CGMH: the iBLEU score of SAGS is higher than that of CGMH by 2--5 points. \red{We also follow the guidelines provided by Carrasco et al.~\cite{carrasco2020recent} to evaluate the performance difference between individual methods by Friedman test and postdoc Nemenyi test. The Friedman test on the comparison between SAGS and the other unsupervised methods presents a chi square statistic of 1083.82 at a p-value lower than $10^{-200}$. We thus rejected the null hypothesis that the configurations are equally performed at the significance level $\alpha=0.05$, and proceed with the post-hoc Nemenyi test analysis. Table~\ref{table:test-text} shows the pairwise comparison results of the Nemenyi test. We observe that the performance of SAGS is clearly different from those of the baselines (P$<$0.05), showing that the optimization improvement obtained by SAGS is not trivial but significant.} These results shows that paraphrase generation is better modeled as an optimization process, instead of sampling from a distribution.

It is curious to see how our unsupervised paraphrase generator is compared with supervised ones, should large-scale parallel data be available. Admittedly, we see that supervised approaches generally outperform SAGS, as they can learn from massive parallel data. Our SAGS nevertheless achieves comparable results with the recent ResidualLSTM model~\citep{prakash2016neural}, reducing the gap between supervised and unsupervised paraphrasing. 

In addition, our SAGS could be easily applied to new datasets and new domains, whereas the supervised setting does not generalize well. This is shown by a domain adaptation experiment, where a supervised model is trained on one domain but tested on the other. 
We notice in Table~\ref{table:wiki} that the performance of supervised models (e.g., Transformer+Copy) decreases drastically on out-of-domain samples, even if both Quora and Wikianswers are  question samples. The performance is supposed to  decrease further if the source and target domains are more different.  SAGS outperforms all supervised domain-adapted paraphrase generators (except DNPG on the Wikianswers dataset).

Table~\ref{table:coco} shows model performance on MSCOCO and Twitter corpora. These datasets are less used for paraphrase generation than Quora and Wikianswers, and thus we could only compare unsupervised approaches by running existing code bases. Again, we see the same trend as Table~\ref{table:wiki}: SAGS achieves the best performance, CGMH second, and VAEs worst. It is also noted that the Twitter corpus yields lower iBLEU scores for all models, largely due to the noise of Twitter utterances~\citep{lan2017a}. However, the consistent results demonstrate that SAGS is robust and generalizable to different domains (without hyperparameter re-tuning).

\begin{table}[!t]
	\begin{center}
	\resizebox{0.7\linewidth}{!}{
		\begin{tabular}{ccccccc}
			\hline\noalign{\smallskip}
		 \multirow{2}*{Model} & \multicolumn{2}{c}{Relevance} &\multicolumn{2}{c}{Fluency}\\
		    \cmidrule(r){2-3}\cmidrule(r){4-5}
			& Mean Score & Agreement  & Mean Score & Agreement \\
			\noalign{\smallskip}
			\hline
			\noalign{\smallskip}
			VAE   & 2.71 & 0.44 & 3.26  & 0.53 \\
			Lag VAE   & 2.86 & 0.47 & 3.27  & 0.50 \\
		    CGMH  & 3.13 & 0.37  & 3.62 & 0.51 \\
			SAGS  & \textbf{3.82} & 0.57 & \textbf{3.67} & 0.52 \\
			\hline
		\end{tabular}
	}
	\end{center}
	\caption{Human evaluation on the Quora dataset.}
	\label{table:human}
\end{table}

\textbf{Human Evaluation.} We also conducted human evaluation on the generated paraphrases. Due to the limit of budget and resources, we sampled 300 samples from the Quora test set and only compared the unsupervised methods (which is the main focus of our work).

We asked six human annotators to evaluate the generated paraphrases in terms of relevance and fluency; each aspect was scored from $1$ to $5$. We report the average human scores and the Cohen's kappa score~\citep{cohen1960coefficient}.
It should be emphasized that our human evaluation was conducted in a blind fashion. Table~\ref{table:human} shows that SAGS achieves the highest human satisfaction scores in terms of both relevance and fluency, and the kappa scores indicate moderate inter-annotator agreement.  The results are also consistent with the automatic metrics in Tables~\ref{table:wiki} and~\ref{table:coco}.  We further conducted two-sided Wilcoxon signed rank tests. The improvement of SAGS is statistically significant with $p<0.01$ in both aspects, compared with both competing methods.

\subsection{Model Analysis}
As our framework involves several novel techniques, here, we take paraphrase generation task as an example to further analyze the design choices of SAGS in more detail. This analysis is conducted on the most widely-used Quora dataset, with a test subset of 2000 samples.

\textbf{Ablation Study.}  We first evaluate the search objective function~(\ref{eq:obj}) in Lines 1--4 of Table~\ref{table:ablation}. The results show that each component of our objective (namely, keyword similarity, sample similarity, and expression diversity) does play its role in  paraphrase generation.

Line~5 of Table~\ref{table:ablation} shows the effect of our copy mechanism, which is used in word replacement and insertion. It yields roughly one iBLEU score improvement if we keep sampling those words in the original sample.

Finally, we test the effect of the temperature decay in SA. Line~6 shows the performance if we fix the initial temperature during the whole search process, which is similar to Metropolis--Hastings sampling.\footnote{The Metropolis--Hastings sampler computes its acceptance rate in a slightly different way from Eqn.~(\ref{eq:temp}).} The result shows the importance of the annealing schedule. It also verifies our intuition that sample generation (in particular, paraphrasing in this paper) should be better modeled as a search problem than a sampling problem.

\begin{figure}[t]
\centering
\includegraphics[width=0.5\linewidth]{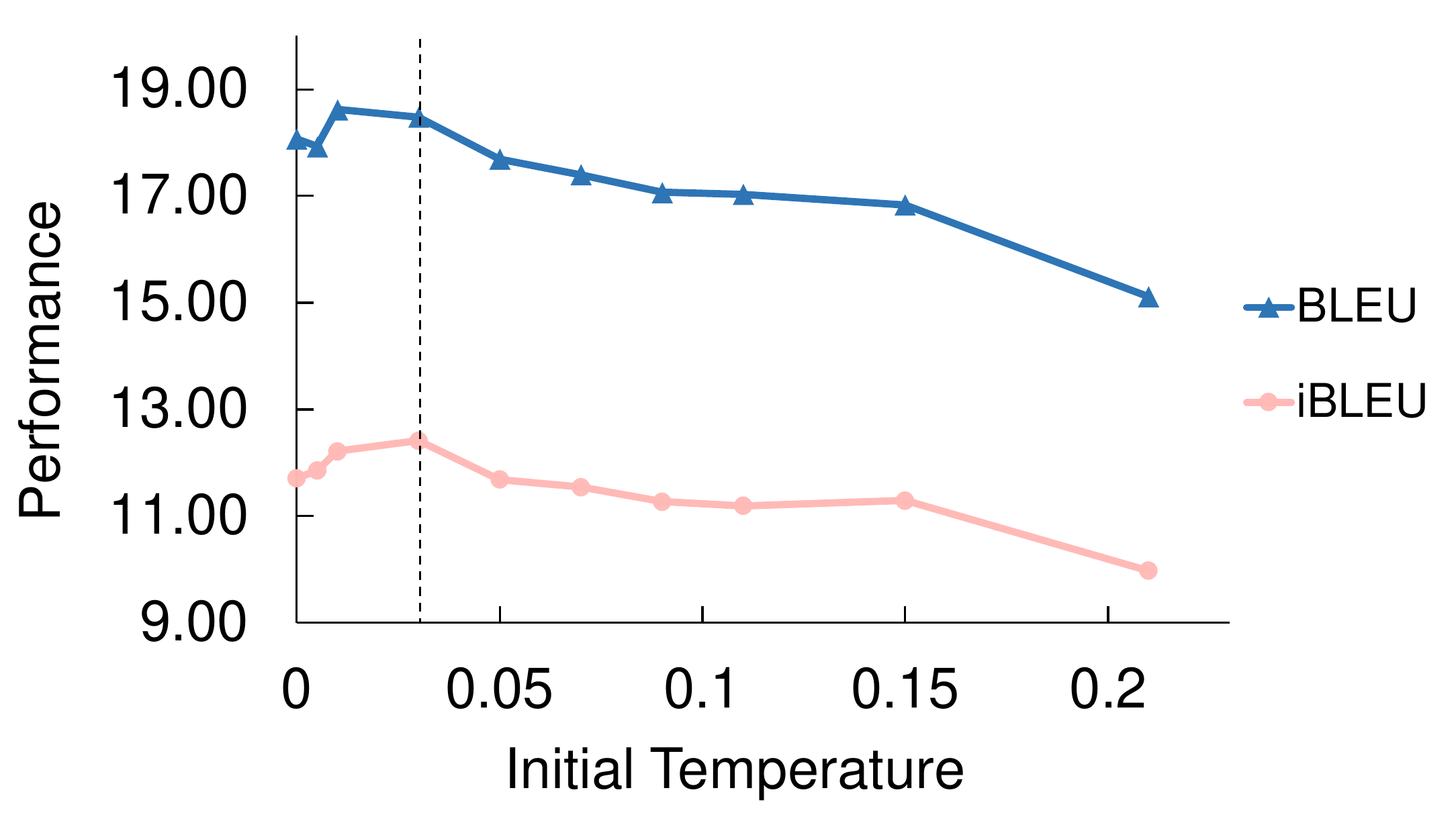}
\caption{Analysis of the initial temperature $T_{\text{init}}$. The dashed line illustrates the selected hyperparameter in validation.}
\label{fig:temp}
\end{figure}

\textbf{Analysis of the Initial Temperature.}  We fixed the decreasing rate to $C=1\times 10^{-4}$ and chose the initial temperature $T_\text{init}$ from $\{0,0.5,1,3,5,7,9,11,15,21\} \times 10^{-2}$. In particular, $T_\text{init}=0$ is equivalent to hill climbing (greedy search). The trend is plotted in Figure~\ref{fig:temp}.

\begin{table}[!t]
	\begin{center}
	\resizebox{0.6\linewidth}{!}{
		\begin{tabular}{clccccc}
			\hline\noalign{\smallskip}
			\!\!\!Line \#\!\!\! & SAGS Variant\!\!\! & iBLEU \!\!\! & BLEU\!\!\! & Rouge1\!\!\!  & Rouge2\!\!\! \\
			\noalign{\smallskip}
			\hline
			\noalign{\smallskip}	
			1 & SAGS   & \textbf{12.41} & 18.48 & 57.06 & 31.39 \\
			\noalign{\smallskip}
			\hline
			\noalign{\smallskip}	
			2& w/o $f_{\text{sim,key}}$ & 10.28  & 15.34  & 50.85 & 26.42\\
			3 & w/o $f_{\text{sim,sen}}$ & 11.78 & 17.95  & 57.04 & 30.80\\
			4  & w/o $f_{\text{exp}}$   & 11.93 & 21.17 & 59.75 & 34.91 \\
			5 & w/o copy  & 11.42 & 17.25 & 56.09 & 29.73\\
			6 & w/o annealing & 10.56 & 16.52  & 56.02 & 29.25 \\
			\hline
		\end{tabular}
		}
	\end{center}
	\caption{Ablation study. }
	\label{table:ablation}
\end{table}

It is seen that a high temperature yields worse performance (with other hyperparameters fixed), because in this case SAGS accepts worse samples and is less likely to settle down. On the other hand, a low temperature makes SAGS greedier, also resulting in worse performance. Especially, our simulated annealing largely outperforms greedy search, whose temperature is 0.

We further observe that BLEU and iBLEU peak at different values of the initial temperature. 
This is because a lower temperature indicates a greedier strategy with less editing, and if the input sample is not changed much, we may indeed have a higher BLEU score. But our major metric iBLEU penalizes the similarity to the input and thus prefers a higher temperature. We chose $T_\text{init}=0.03$ by validating on iBLEU.

\textbf{Analysis of Cooling Schedules.} As the above experiments show that the temperature can significantly affect the search performance of SAGS, we are going to investigate the optimization performances of different cooling schedules in SAGS. The performances of the individual SA variants with the above different cooling schedules are shown in Table~\ref{table:cooling-para}. Table~\ref{table:cooling-para} also presents the average convergence steps of each cooling schedule. We observed that the three cooling schedules yield similar results, the linear cooling schedule performs slightly better, which may be attributed to the more editing iterations of convergence of the linear cooling schedule. Overall, the SAGS variants with different cooling schedules yield similar performances. Therefore, after the hyperparameter tuning via grid searches, the choice of the cooling schedule in SAGS does not significantly influence the optimization performance. More experimental details can be found in Appendix A.

\begin{table*}[t]\centering\footnotesize
	\resizebox{\linewidth}{!}{
		\begin{tabular}{|p{0.18\linewidth}|p{0.22\linewidth}|p{0.195\linewidth}|p{0.215\linewidth}|p{0.22\linewidth}|}
		\hline
		\multicolumn{1}{|c}{Input} & \multicolumn{1}{|c}{VAE} & \multicolumn{1}{|c}{Lag VAE}    & \multicolumn{1}{|c|}{CGMH}   & \multicolumn{1}{|c|}{SAGS} \\
		\hline
		where are best places for spring snowboarding in the us? & where are best places for running in the world? (3.33)  & where are best places for honeymoon year near the us? (2.33) & where is best store for the snowboarding in the US? (3.67) & where can I find the best places in the US for snowboarding? (4.67) \\\hline
		how can i become good in studies? & how can i have a good android phone? (2.33)  & how can i become good students? (4.33)& how can i become very rich in studies? (4.00) & how should i do to get better grades in my studies? (4.33) \\\hline
		what are the pluses and minuses about life as a foreigner in singapore? & what are the UNK and  most interesting life as a foreigner in medieval greece? (2.33) &  what are the UNK and interesting things about life as a foreigner? (2.33) & what are the misconception about UNK with life as a foreigner in western? (2.33) & what are the mistakes and pluses life as a foreigner in singapore?  (2.67)  \\\hline
		\end{tabular}
		}
		\caption{Example paraphrases generated by different methods on the Quora dataset. The averaged score evaluated by three annotators is shown at the end of each generated sample. }
	\label{table:example}
\end{table*}

\textbf{Case Study.}
We showcase several generated paraphrases in Table~\ref{table:example}. We see qualitatively that SAGS can produce more reasonable paraphrases than the other methods in terms of both closeness in meaning and difference in expressions, and can make non-local transformations. For example, ``\textit{places for spring snowboarding in the US}'' is paraphrased as ``\textit{places in the US for snowboarding}.'' Admittedly, such samples are relatively rare, and our current SAGS mainly synthesizes paraphrases by editing  words in the sample, whereas the syntax 
is mostly preserved. This is partially due to the difficulty of exploring the entire (discrete) sample space even by simulated annealing, and partially due to the insensitivity of the similarity objective  given two very different samples.

\section{Experiments on Molecule Generation}
Besides the sequence optimization, SAGS can also work on graphs. We take molecular generation as an example to show how to apply SAGS to perform graph optimization. We first follow Jin et al.~\cite{junctiontree} and Shi et al.~\cite{2020graphaf} to optimize the molecular solubility for 800 molecules in ZINC250K~\cite{irwin2012zinc} with the lowest scores. Besides the single property optimization, we then conduct multiple-objective molecular optimization.

\subsection{Competing Methods and Implementation Details} 
In recent years, numerous efforts have been devoted to molecular optimization and noticeable development was achieved. The related works include JT-VAE, GCPN, matched molecular pair analysis (MMPA), \red{JADE}, MoFlow, GraphAF, \red{CMA-ES, the coyote algorithm (COA) and the molecule swarm optimization (MSO)}. JT-VAE~\cite{junctiontree} generates molecules by first decoding a tree structure of chemical groups and then assembling them into molecules. GCPN~\cite{gcpn} learns to optimize domain-specific rewards through policy gradient. MMPA accomplishes molecule graph optimization based on hard-coded  graph transformation rules.  \red{JADE~\cite{zhang2009jade} is an adaptive differential evolution algorithm that adopts a new mutation strategy and updates the control parameters in an adaptive manner. MoFlow~\cite{zang2020moflow} is a discrete latent variable model for molecular graph generation based on normalizing flow methods. COA~\cite{pierezan2018coyote} is a population based metaheuristic for optimization inspired on the canis latrans species. CMA-ES (the covariance matrix adaptation evolution strategy)~\cite{hansen1996adapting} is an evolutionary algorithm for difficult non-linear non-convex black-box optimization problems in continuous domain.} GraphAF~\cite{2020graphaf} is a flow-based autoregressive model that generates the nodes and edges based on existing sub-graphs. MSO is a PSO-based molecule generation algorithm that works on a learned continuous latent space~\cite{winter2019efficient}. We take all of these methods into comparison.

Similar to the experiments on sequences, we also apply a grid search to determine the hyperparameters of the graph-based SAGS on the validation set of the ZINC250K dataset. The validation set contains 200 molecules that are different from the test set (i.e., the 800 molecules with lowest penalized logP scores). The hyperparameters of the graph-based SAGS involves the initial temperature $T_\text{init}$, the cooling coefficient $C$ and the weight in the objective $M$. Due to the limitation of computational resources, we are not able to search all the possible combinations of these hyperparameters. Instead, we only empirically consider a coarse search range. Specifically, the initial temperature $T_\text{init}$ is chosen from \{1, 3, 5, 7, 9\}$\times 10^{-2}$ and set to 0.01; The cooling coefficient $C$ is chosen from \{1, 3, 5, 7, 9\}$\times 10^{-6}$ and set to $3\times10^{-6}$; The weight $M$ in the objective was chosen from \{0.5, 1, 3, 5, 7, 9\} and set to 5. The search range of $M$ is different from the power weight $P$ because the scoring functions of molecule generation are additive. We also conduct a similar hyperparameter tuning process for SAGS (GA). Details are not repeated.

 \begin{table*}[t]	
 	\begin{center}
 	\small
 	\resizebox{0.9\linewidth}{!}{
 		\begin{tabular}{lccccccccccccccccc}
 			\hline\noalign{\smallskip}
 		    \multirow{2}*{Methods}& \multicolumn{2}{c}{Similarity $\ge$ 0.0} &\multicolumn{2}{c}{Similarity $\ge$ 0.2}  &\multicolumn{2}{c}{Similarity $\ge$ 0.4} &\multicolumn{2}{c}{Similarity $\ge$ 0.6}\\
 		   \cmidrule(r){2-3}\cmidrule(r){4-5}\cmidrule(r){6-7} \cmidrule(r){8-9}
 		      &  Improvement  & Success &  Improvement   & Success & Improvement   & Success  & Improvement   & Success \\
 			\hline
 			\noalign{\smallskip}
 			JTVAE  &  1.91 $\pm$ 2.04 & 97.50\% & 	1.68 $\pm$  1.85 & 	 97.10\% & 0.84 $\pm$ 1.45  & 83.60\% & 0.21 $\pm$ 0.71 &  46.40\% \\
 			GCPN  &  4.20 $\pm$ 1.28 &  100\% & 	4.12 $\pm$  1.19 & 	 100\% & 2.49 $\pm$ 1.30 &  100\% & 0.79 $\pm$ 0.63 & 100\% \\
             JADE  &  8.55 $\pm$ 7.46 & 90.13\% & 	4.11 $\pm$  6.19 & 57.75\% & 2.45 $\pm$ 5.72 & 50.25\% & 0.15 $\pm$ 5.06  & 34.88\% \\
 			MMPA  & - & - & - & -  & 3.29 $\pm$ 1.12  & - & 1.65 $\pm$ 1.44 &  - \\
   			 MoFlow  & 8.61 $\pm$ 5.44 & 98.88\% & 7.06 $\pm$ 5.04 & 96.75\%& 4.71 $\pm$ 4.55  & 85.75\%& 2.10 $\pm$ 2.86&  58.25\% \\
             GraphAF  &  13.13 $\pm$ 6.89 & 100\% & 	11.90 $\pm$  6.86 & 	 100\% & 8.21 $\pm$ 6.51 & 99.88\% & 4.98 $\pm$ 6.49  & 96.88\% \\
             CMA-ES  & 13.35 $\pm$ 6.49 & 100\% & 11.62 $\pm$ 7.43 & 97.75\% & 6.86 $\pm$ 9.33 & 	67.38\% &	3.04 $\pm$9.65	& 34.25\% \\
             COA  &  13.76 $\pm$ 6.19 & 100\% & 	11.09 $\pm$ 7.32  & 96.88\% & 7.32 $\pm$ 7.31 & 78.75\% & 6.14 $\pm$ 7.05  & 78.50\% \\
             MSO  &  22.77 $\pm$ 6.31 & 100\% & 	18.22 $\pm$  6.23 & 99.75\% & 12.11 $\pm$ 8.37 & 84.25\% & 6.55 $\pm$ 8.47  & 52.5\% \\
             SAGS (GA)  &  \textbf{27.53 $\pm$ 19.29} & 100\% & 	\textbf{19.42 $\pm$  17.70} & 	 96.75\% & 10.05$\pm$ 11.10 & 70.63\% &4.86$\pm$ 7.02  & 34.88\% \\
     		 SAGS & {17.38 $\pm$ 6.76}* & 99.88\% & {15.07$\pm$ 6.98}* &  {96.25\%}  & \textbf{{12.73 $\pm$ 6.59}}*  & {85.88\%} &  \textbf{{8.83 $\pm$6.02}}*  & {83.25\%}\\
 			\noalign{\smallskip}
 			\hline
 		\end{tabular}
 	}
 	\end{center}
 		\caption{Performance on the optimization of Penalized LogP property. The dash sign indicates the results of the corresponding methods are not available. \red{*: P$<$0.01 by the Friedman and postdoc Nemenyi test in the comparison between SAGS and JADE, CMA-ES, COA, MSO and SAGS (GA), respectively. See Table~\ref{table:test-mol} for more statistical results.}}
 	\label{table:logp}
 \end{table*}

\begin{table}[t]
 \begin{center}
	\small
	\resizebox{0.45\linewidth}{!}{
		\begin{tabular}{ccccccccccccccc}
		\hline
		    \noalign{\smallskip}
		   Similarity threshold ($\delta$) & SAGS & Annotated label \\
			\noalign{\smallskip}
		    \hline
			\noalign{\smallskip}
			 0.4 & 4.27 $\pm$ 3.27  & 3.76 $\pm$ 1.26 \\
			 0.6 &	1.70 $\pm$ 2.26 & 1.89 $\pm$ 0.72 \\
			\hline		
		\end{tabular}
		}
		\end{center}
	\caption{Comparison in terms of the optimized penalized LogP profiles between graph-based SAGS and the annotated labels in two parallel datasets (similarity thresholds of 0.4 and 0.6, other thresholds are not available). }
	\label{table:upper}
\end{table}

\subsection{Molecule Generation of Single-Objective Optimization}
In the task of single-objective optimization, we adopt four different levels of the similarity constraints, namely, the similarity above 0.0, 0.2, 0.4 and 0.6, respectively. That is, we compare the optimization performance of individual methods on these four conditions.
As shown in Table \ref{table:logp},  JT-VAE obtains the worse optimization performance, \red{showing the difficulty to learn a robust latent space of graphs. RL-based methods, such as GCPN and GraphAF, present better optimization results, but still lag behind SAGS by a large margin. The advanced meta-heuristics, such as CMA-ES, COA, MSO and the genetic algorithms, further boost the optimization performance. Compared with SAGS,} SAGS (GA) and MSO obtain lower scores when the similarity threshold is higher than 0.4 and perform better in the other settings. In this task, we mainly focus on the strong similarity constraints (high similarity thresholds) since they mimic the realistic demands of preserving some of the key functional groups. Note that SAGS (GA) works better on molecule generation than the paraphrase generation task \red{because} the crossover operation can make more meaningful candidate samples of molecules. Considering that the crossover operation performs global edit of the molecular structure, GA presents poorer optimization performance when the similarity constraint is strong. Overall, the above results clearly indicate the optimization effectiveness of the graph-based SAGS.

To further study the optimization power of the graph-based SAGS, we collect two parallel datasets that contain \red{800} paired molecules (X,Y) from Jin et al.~\cite{jin2019hierarchical}, with similarity thresholds of 0.4 and 0.6, respectively. These molecule pairs are conventionally used to guide a model to translate a source molecule X to a target molecule Y with higher penalized logP property. Therefore, these parallel molecule pairs provide an opportunity to approximate the upper bound of the optimization performance that a machine learning method can reach. We feed the source molecule X into graph-based SAGS and compare the quality of the outputs of SAGS with the annotated label Y. As shown in Table \ref{table:upper}, SAGS presents comparable results to the annotated labels of these parallel datasets. Considering SAGS is an unsupervised method, these results further demonstrate the sufficient search capacity of our approach.

\begin{figure}[!t]
	\centering
	\includegraphics[width=0.8\linewidth]{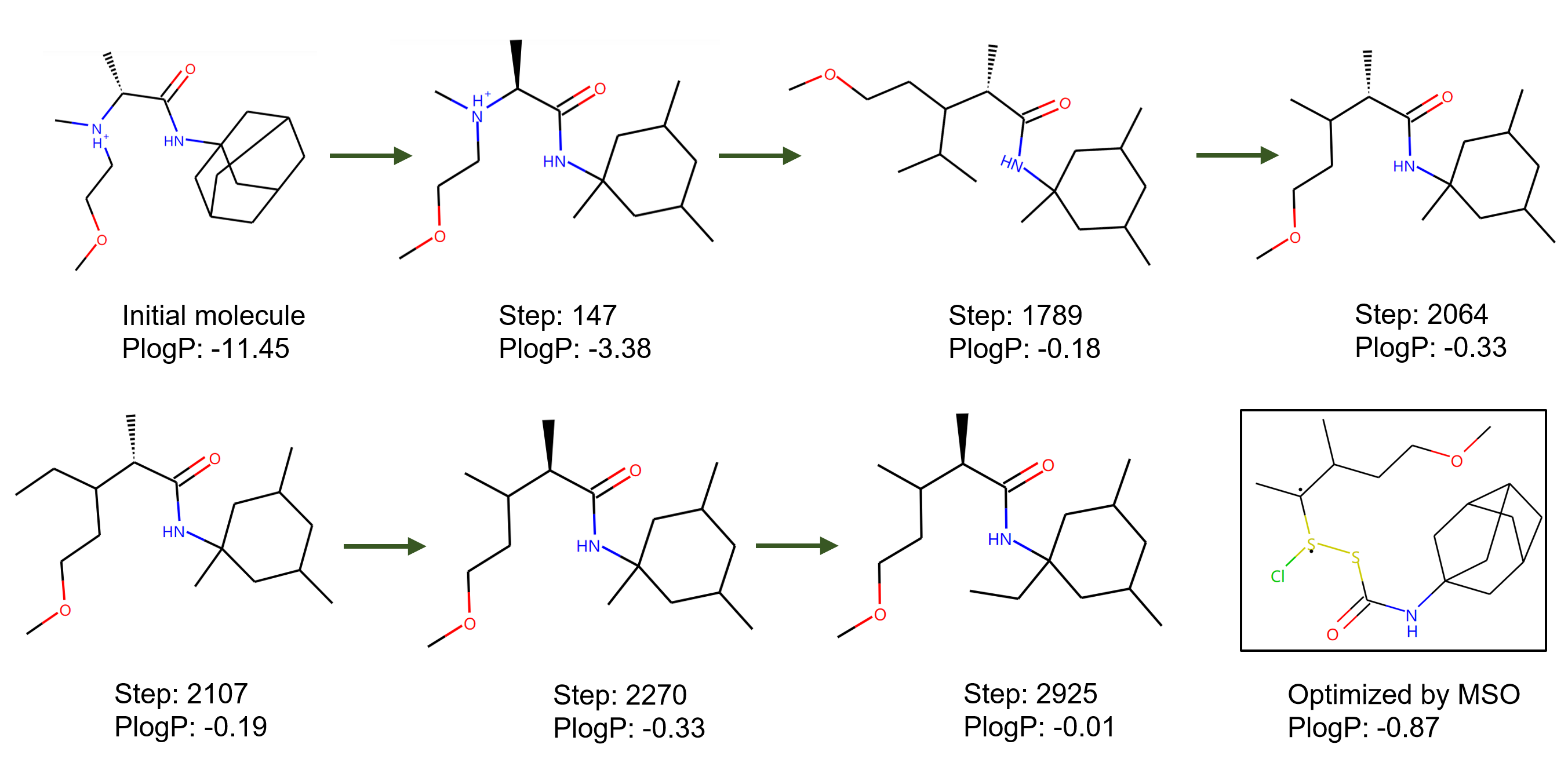}
	\caption{Examples of the optimization process SAGS for molecule generation. Only accepted modifications are shown. The similarity threshold of 0.4 is used in this experiment.
	}
	\label{fig:case_mol}
\end{figure}

\textbf{Case Study.} 
To have an intuitive understanding of the sampling strategy of SAGS, we randomly choose a molecule and illustrate the optimization process in Figure \ref{fig:case_mol}. This molecule is an endocyclic compound with several hydrophilic groups, such as the cationic amine, and thus shows a low penalized logP score (denoted as PlogP). SAGS firstly SAGS removes the bridge connection in the ring, which largely increases the PlogP. Then SAGS replaces the cationic amine with a neutral amine to improve the hydrophobicity. Afterward, SAGS searches to extend the carbon chain to further improve the hydrophobicity, finally resulting in an improvement of 11.44 in terms of PlogP.

In addition, Figure~\ref{fig:case_mol} also illustrates the optimization results of MSO that uses the same molecule as input. Compared with SAGS, we noticed that the PlogP score of the molecule generated by MSO is slightly lower. Also, the molecule contains carbon and sulfur free radical, which is not stable and difficult to synthesize. By contrast, as SAGS employs the pre-trained molecular graph model to provide reasonable molecule proposals, the generated molecule may be more preferred in realistic applications.

\subsection{Molecule Generation of Multi-Objective Optimization}
Next, we tried to apply our framework to the multi-objective optimization task. In the multi-objective optimization task, we aim to maximize both PlogP and QED scores under a predefined similarity threshold (i.e., $\delta$=0.6).  We compare SAGS with the competing baseline MSO using the identical objective function. The objective function is defined by the geometric average score between PlogP (normalized by the maximum one in the dataset) and QED. In this experiment, the test data were a total of 400 molecules, all of which were randomly selected from the ChEMBL dataset. ChEMBL is a large-scale dataset that contains more than 1 million bioactive drug-like small molecules. The performance of MSO was obtained by running its released source code. Table~\ref{table:multip} shows the comparison results between SAGS and MSO, where SAGS significantly outperforms MSO in terms of the improvement of both the PlogP and QED scores. We noticed that SAGS obtained a higher improvement of PlogP on ChEMBL than that on ZINK250K, this is because the PlogP scores of the input molecules on ChEMBL are much lower (the average PlogP score is -29.48). This result further establishes the search effectiveness of SAGS on graph optimization.

\begin{table}[!t]
	\begin{center}
	\resizebox{0.35\linewidth}{!}{
		\begin{tabular}{ccccccc}
			\hline\noalign{\smallskip}
		 \multirow{2}*{Method} & \multicolumn{2}{c}{PlogP} &\multicolumn{2}{c}{QED}\\
		    \cmidrule(r){2-3}\cmidrule(r){4-5}
			& Imp. & Succ.  & Imp. & Succ. \\
			\noalign{\smallskip}
			\hline
			\noalign{\smallskip}
			MSO   & 4.75 & 100\% & 0.05  & 100\% \\
			SAGS  & \textbf{15.37} & 100\% & \textbf{0.16} & 100\% \\
			\hline
		\end{tabular}
	}
	\end{center}
	\caption{Comparison between SAGS and MSO on the multi-objective optimization task on the test molecules from the ChEMBL dataset.}
	\label{table:multip}
\end{table}

\section{Conclusion}
In this paper, we proposed a novel unsupervised approach, SAGS, that can optimize graphs and sequences by simulated annealing. SAGS provides a new discrete optimization algorithm that integrates the deep generative model into edit-based simulated annealing sampling. We propose a search objective function, considering both the property of the structure and grammar constraints. Experiments on five benchmark datasets across different domains show that our model outperforms previous state-of-the-art unsupervised methods to a large extent. In the paraphrase generation task, we further surpass most domain-adaptive paraphrase generators, as well as a supervised model on the Wikianswers dataset. In molecule generation, SAGS also shows the optimization superiority over existing methods in both single and multi-objective optimization tasks. \red{Besides the above two problems, our framework also has the potential to address a wide variety of sequential pattern mining problems~\cite{fournier2017survey} (e.g., text summarization and text simplification). For example, given a score function that evaluates the quality of the candidate summarization, SAGS can generate meaningful summarization sentences by iteratively editing the original text.
}

In the future, we plan to further equip the SAGS framework with the operations that work globally. For example, we plan to impose syntactic parse trees in hopes of generating more syntactically different paraphrases and leverage the operation on the functional groups for molecule generation.

\section*{Acknowledgments}

This work was supported in part by funds from the key scientific technological innovation research project by Ministry of Education and National Natural Science Foundation of China (61836004). We also acknowledge the support of Beijing Brain Science Special Project (No.Z181100001518006) and a grant from Institute Guo Qiang, Tsinghua University. This work is also partially supported by Compute Canada (www.computecanada.ca), the Alberta Machine Intelligence Institute (Amii) Fellow Program, the Canadian CIFAR AI Chair Program, and the Natural Sciences and Engineering Research Council of Canada (NSERC) under Grant No. RGPIN-2020-04465.

\bibliographystyle{apalike}
\bibliography{ref}

\appendix
\setcounter{table}{0}

\section{Experimental Details of the Analysis of the Cooling Schedules}
Here, we consider an exponential cooling schedule and also a logarithmic one. we use the following form of the exponential cooling schedule,
\begin{align}
T= T_{\text{init}}\cdot e^{C_et}
\end{align}
where $T_{init}$ and $C_e$ are the hyperparameters to control the annealing process. $C_e$ is the exponential cooling coefficient. The logarithmic cooling schedule we used is
\begin{align}
T=\frac{T_{\text{init}}}{log(e+C_l\cdot t)}
\end{align}
where $C_l$ is the logarithmic cooling coefficient.

We first apply these two cooling strategies to the paraphrase generation task. To fairly compare the individual SA variants with the above different cooling schedules, we apply a grid search similar to the linear cooling schedule to calibrate the hyperparameters. In particular,  $C_e$ and $C_l$ are chosen from \{0.05,0.1,0.2,0.4.0.8\} on the validation dataset. The search range of $T_{\text{init}}$ is the same as that in the linear cooling schedule. Finally, in the exponential cooling schedule,  $C_e$is set to 0.05 and  $T_{\text{init}}$ is set to 0.03; In the logarithmic cooling schedule, $C_l$ is set to 0.8 and $T_{\text{init}}$is set to 0.05. 

The performances of the individual SA variants with the above different cooling schedules are shown in Table~\ref{table:cooling-para}. Table~\ref{table:cooling-para} also presents the average convergence steps of each cooling schedule. We observed that the three cooling schedules yield similar results, the linear cooling strategy performs slightly better.

Then, we study these cooling schedules on the molecule generation task. In this experiment, we used the 800 molecules with the lowest PlogP scores in the ZINC250K dataset and the similarity threshold is set to 0.6. The best hyperparameters of the individual cooling schedules are also calibrated by a similar grid search. Finally, in the exponential cooling schedule, $C_e$ is set to 0.1 and  $T_{\text{init}}$ is set to 0.01; In the logarithmic cooling schedule, $C_l$ is set to 0.1 and  $T_{\text{init}}$ is set to 0.1. The comparison of the cooling schedules on molecule generation is shown in Table~\ref{table:cooling-mol}. Similar to the paraphrasing task, all the cooling schedules achieve similar superior optimization results. The logarithmic schedule yields slightly better results in molecule generation. Also, the number of convergence steps of these cooling schedules is also similar. Therefore, after the hyperparameter tuning via grid searches, the choice of the cooling schedule in SAGS does not significantly influence the optimization performance.

\section{Supplementary Tables}

\begin{table}[H]
	\centering
	\tiny
	\resizebox{0.75\linewidth}{!}{
		\begin{tabular}{ccccccccc}
			\hline
			\multirow{2}{*}{Initial temperature}  & \multicolumn{6}{c}{Cooling coefficient}  \\
			\cmidrule(r){2-7}
			 & 0.5$\times$10\textsuperscript{\tiny{-4}}& 1$\times$10\textsuperscript{\tiny{-4}} & 3$\times$10\textsuperscript{\tiny{-4}} & 5$\times$10\textsuperscript{\tiny{-4}} & 7$\times$10\textsuperscript{\tiny{-4}} & 9$\times$10\textsuperscript{\tiny{-4}} \\
			\hline
			0.005 & 12.21 &	11.94	&11.85&	11.69&	11.54	&11.63 \\
			0.01  & 12.07&	12.23	&12.21&	11.96&	11.77	&11.72 \\
			0.03  & 11.87&	12.03	&\textbf{12.38}&	12.19&	11.97	&11.83 \\
			0.05  & 11.43&	11.61	&11.68&	12.35&	12.14	&11.97 \\
			0.07  & 11.29&	11.38	&11.54&	11.51&	12.29	&12.01 \\
			0.09  & 11.18&	11.03	&11.27&	11.11&	12.02	&12.23 \\
			\hline
		\end{tabular}
	}
	\caption{The paraphrasing performances of SAGS with different combinations of the initial temperature $T_\text{init}$ and the cooling coefficient $C$ in terms of the iBLEU score. The best performance is highlighted in bold. This experiment was conducted on the validation set of the Quora dataset, with the other hyperparameters fixed (i.e., the three power weights in the objective function $Q=1,S=1,P=8$).}
	\label{tab:hyper-t}
\end{table}

\begin{table}[H]
	\centering
	\tiny
	\resizebox{0.7\linewidth}{!}{
		\begin{tabular}{ccccccccc}
			\hline
			\multirow{2}{*}{Power weight of expression diversity}  & \multicolumn{5}{c}{Power weight of keyword similarity}  \\
			\cmidrule(r){2-6}
			 & 0.5 & 1 & 2 & 4 & 8 \\
			\hline
			0.5& 12.13	&12.19&	12.24&	12.29 &	12.36 \\
			1  & 11.38	&11.84&	12.05&	12.34 &	\textbf{12.38} \\
			2  & 10.46	&10.67&	10.82&	11.09 &	11.18 \\
            4  & 9.12	&9.29	&9.89&	10.28 &	10.42 \\
            8  & 8.08	&8.38	&8.87&	9.03 & 9.51  \\
			\hline
		\end{tabular}
	}
	\caption{The paraphrasing performances of SAGS with different combinations of the two power weights in the objective function (i.e., the power weight of the expression diversity $S$ and the power weight of the keyword similarity $P$)  in terms of the iBLEU score. The best performance is highlighted in bold. This experiment was conducted on the validation set of the Quora dataset, with the other hyperparameters fixed (i.e., the power weight of semantic similarity $Q=1$, the initial temperature $T_\text{init}=0.03$ and the cooling coefficient $C=3\times 10^{-4}$).}
	\label{tab:hyper-2}
\end{table}

\begin{table}[H]
 \begin{center}
	\small
	\resizebox{0.5\linewidth}{!}{
		\begin{tabular}{ccccccccccccccc}
		\hline
		    \noalign{\smallskip}
		   Methods & VAE &	Lag VAE &	CGMH &	SAGS (GA)	& SAGS \\
			\noalign{\smallskip}
		    \hline
			\noalign{\smallskip}
		    VAE & 1 & 0.001 & 0.001 & 0.001 & 0.001 \\
			Lag VAE &  0.001 & 1 & 0.001 & 0.001 & 0.001 \\
			CGMH & 0.001 & 0.001 & 1 & 0.001 & 0.001 \\
			SAGS (GA) & 0.001 & 0.001 & 0.001 & 1 & 0.001\\
			SAGS & 0.001 & 0.001 & 0.001 & 0.001 & 1
			\\
			\hline		
		\end{tabular}
		}
		\end{center}
	\caption{\red{The p-values of Nemenyi pairwise comparison between the paraphrasing methods on the Quora dataset. }}
	\label{table:test-text}
\end{table}

\begin{table}[H]
 \begin{center}
	\small
	\resizebox{0.6\linewidth}{!}{
		\begin{tabular}{ccccccccccccccc}
		\hline
		    \noalign{\smallskip}
		   Cooling schedule & iBLEU	& BLEU	& Rouge1 &	Rouge2 &	Convergence steps \\
			\noalign{\smallskip}
		    \hline
			\noalign{\smallskip}
		Exponential &     12.13&18.33&	56.53&	31.02	&58.34 \\
			 Logarithmic& 12.03&17.98&	55.42&	30.18	&63.91\\
			 Linear &     12.41&18.48&	57.06&	31.39	&89.53 \\
			\hline		
		\end{tabular}
		}
		\end{center}
	\caption{Comparison of the different cooling strategies in paraphrase generation on the Quora dataset. The test sentences are totally 2000, all of which are randomly sampled from the test set of Quora.}
	\label{table:cooling-para}
\end{table}

\begin{table}[H]
 \begin{center}
	\small
	\resizebox{0.5\linewidth}{!}{
		\begin{tabular}{ccccccccccccccc}
		\hline
		    \noalign{\smallskip}
		   Methods & JADE &	CMA-ES	& COA	& MSO &	SAGS (GA)	& SAGS \\
			\noalign{\smallskip}
		    \hline
			\noalign{\smallskip}
		    JADE & 1 & 0.001 & 0.001 & 0.001  &0.001 & 0.001 \\
			CMA-ES &  0.001 & 1 & 0.001 & 0.001 &0.001  & 0.001 \\
			COA & 0.001 & 0.001 & 1 & 0.001 & 0.001  & 0.001 \\
			MSO & 0.001 & 0.001 & 0.001  & 1 & 0.001 & 0.001   \\
			SAGS (GA) & 0.001 & 0.001 &  0.001  & 0.001 & 1 & 0.001\\
			SAGS & 0.001 & 0.001 & 0.001  & 0.001  & 0.001 & 1
			\\
			\hline		
		\end{tabular}
		}
		\end{center}
	\caption{\red{The p-values of the Nemenyi pairwise comparison between the molecule generation methods on the ZINK250K dataset. } }
	\label{table:test-mol}
\end{table}

\begin{table}[H]
 \begin{center}
	\small
	\resizebox{0.45\linewidth}{!}{
		\begin{tabular}{ccccccccccccccc}
		\hline
		    \noalign{\smallskip}
		   Cooling schedule & Imp. & Succ. & Convergence steps \\
			\noalign{\smallskip}
		    \hline
			\noalign{\smallskip}
		Exponential & 8.80  & 100\% & 2984.03 \\
			 Logarithmic& 9.02  &  100\%  & 2987.13 \\
			 Linear & 8.88 &  100\%  &  2987.07 \\
			\hline		
		\end{tabular}
		}
		\end{center}
	\caption{Comparison of the different cooling strategies in molecule generation. The similarity threshold of 0.6 is used in this experiment. }
	\label{table:cooling-mol}
\end{table}

\end{document}